\documentclass[a4paper,fleqn]{cas-dc}

\usepackage[numbers,sort,compress]{natbib}
\def\tsc#1{\csdef{#1}{\textsc{\lowercase{#1}}\xspace}}
\tsc{WGM}
\tsc{QE}
\usepackage{pifont}
\usepackage{multirow}
\usepackage{makecell}
\usepackage{stfloats}
\usepackage{bbding}
\usepackage{amsmath}
\usepackage{algorithmic}
\usepackage{algorithm,algorithmic}
\usepackage{graphics}
\usepackage{multirow}
\usepackage[switch]{lineno}

\begin{document}
\let\WriteBookmarks\relax
\def\floatpagepagefraction{1}
\def\textpagefraction{.001}
\let\printorcid\relax
%\linenumbers

\shorttitle{Opportunities and challenges in the application of large artificial intelligence models in radiology}    

\shortauthors{L. Pan et al.}  

\title [mode = title]{Opportunities and challenges in the application of large artificial intelligence models in radiology}  

\author[a]{Liangrui Pan} \ead{panlr@hnu.edu.cn}
\author[b]{Zhenyu Zhao} \ead{188202114@csu.edu.cn}
\author[c]{Ying Lu} \ead{luying@nudt.edu.cn }
\author[d]{Kewei Tang} \ead{kewei.tang@fullink.ai}
\author[e]{Liyong Fu}\cormark[1] \ead{fuliyong840909@163.com}
\author[f]{Qingchun Liang}\cormark[1] \ead{503079@csu.edu.cn}

%\author[a]{Wenjuan Liu}  \ead{liuwenjuan89@hnu.edu}
%\author[a]{Liwen Xu} \cormark[1]\ead{xuliwen@hnu.edu.cn}
\author[a]{Shaoliang Peng}\cormark[1]\ead{slpeng@hnu.edu.cn}

\affiliation[a]{organization={College of Computer Science and Electronic 
							  Engineering, Hunan University},
				city={Changsha},
				postcode={410083}, 
				state={Hunan},
				country={China}}

\affiliation[b]{organization={Department of Thoracic Surgery, The second xiangya hospital, National University of Defense Technology},
	city={Changsha},
	postcode={410000}, 
	state={Hunan},
	country={China}}

\affiliation[c]{organization={College of Military and Political Basic Education, National University of Defense Technology},
	city={Changsha},
	postcode={410072}, 
	state={Hunan},
	country={China}}

\affiliation[d]{organization={Fullink Technology Group},
	city={Hangzhou},
	postcode={310000}, 
	state={Zhejiang},
	country={China}}

\affiliation[e]{organization={Institute of Forest Resource Information Techniques, Chinese Academy of Forestry},
	city={Beijing},
	postcode={100091}, 
	country={China}}
	
\affiliation[f]{organization={Department of Pathology, The second xiangya hospital, Central South University},
		city={Changsha},
		postcode={410000}, 
		state={Hunan},
		country={China}}
		
%\affiliation[d]{organization={National Supercomputing Center in Zhengzhou, Zhengzhou University},
%	city={Zhengzhou},
%	postcode={450001}, 
%	state={Henan},
%	country={China}}		

\cortext[1]{Corresponding author}
%\cortext[2]{Corresponding author}

\begin{abstract}
	Influenced by ChatGPT, artificial intelligence (AI) large models have witnessed a global upsurge in large model research and development. As people enjoy the convenience by this AI large model, more and more large models in subdivided fields are gradually being proposed, especially large models in radiology imaging field. This article first introduces the development history of large models, technical details, workflow, working principles of multimodal large models and working principles of video generation large models. Secondly, we summarize the latest research progress of AI large models in radiology education, radiology report generation, applications of unimodal and multimodal radiology. Finally, this paper also summarizes some of the challenges of large AI models in radiology, with the aim of better promoting the rapid revolution in the field of radiography.

\end{abstract}


\maketitle

% Main text
\section{Introduction}

In late 2022, OpenAI unveiled an artificial intelligence chat program named ChatGPT (Chat generative pre-trained transformer), garnering widespread attention across various industries \cite{Roumeliotis2023-dl}. This marks the inaugural demonstration of a large AI model capable of handling diverse open tasks on a global scale. ChatGPT, a natural language processing technology grounded in artificial intelligence, produces responses aligned with language conventions and logic, drawing from provided questions and context \cite{Ray2023-qa,malik2023so}. This technology finds applications in diverse fields including customer service, intelligent assistants, education, and healthcare, facilitating convenient and efficient access to information for individuals \cite{Ray2023-qa,malik2023so}. Presently, AI large models find application across multiple domains, with a growing prevalence in natural language processing, image and text generation, among others, garnering widespread recognition and user appreciation \cite{Navigli2023-ap}. As technology continues to advance, numerous domestic tech companies have introduced their large language model products, including Baidu's knowledge-enhanced large-scale language model, Wenxin Yiyan \cite{Sun2021-bc}, ByteDance's Skylark, and iFlytek's Spark \cite{liu2023research}. Hence, the rising popularity of large AI models stems from their ability to deliver efficient, intelligent services and their ongoing innovations to better fulfill people's needs \cite{Navigli2023-ap}.

ChatGPT is a substantial pre-training model in natural language processing. It employs a deep neural network with numerous parameters, trains it on extensive unlabeled data, and subsequently fine-tunes the large pre-training model for downstream tasks \cite{Kalyan2024-kd,Min2024-nc}. The model can excel in specific tasks, demonstrating outstanding performance. Recently, there has been a surge in interest regarding the integration of medical imaging and AI large models. Certain practical applications of AI large models have captured the interest of medical educators and professionals. The advent of ChatGPT will present both new opportunities and challenges for the advancement of education and medicine. This article initially presents the developmental history, technical intricacies, workflow, and operational principles of multimodal and video generation large models. Subsequently, this article delves into the detailed discussion of AI large models' application in radiology, encompassing education, report generation, and both unimodal and multimodal applications. This article aims to serve as a reference for promoting the establishment of an "radiation medicine + artificial intelligence" education system.

\section{Artificial intelligence large model technology and progress}
\subsection{AI large model development history}
In 2020, OpenAI first proposed the "law of scale," suggesting that model performance will linearly enhance with the exponential expansion of parameter volume, data volume, and training time, with minimal reliance on architecture and optimized hyperparameters \cite{Kaplan2020-sr,Miller2024-yt}. Researchers have shifted their focus towards large models. The technological advancement of the ChatGPT series models epitomizes the progress of large models \cite{Cheng2023-ah}. GPT-1 primarily adopts the architecture of a generative, decoder-only transformer, employing a hybrid approach of unsupervised training and supervised fine-tuning \cite{Zhao2023-ku}. Although GPT-1 has excelled in numerous natural language processing tasks, it encounters challenges in generating lengthy texts while maintaining contextual coherence. GPT-2 employs a structure akin to GPT-1, yet with a parameter size reaching 1.5 billion, trained on the extensive WebText dataset \cite{Thirunavukarasu2023-vc}. Primarily trained through unsupervised methods, it excels in generating more extensive and coherent text. GPT-3 boasts a model parameter of 175 billion, demonstrating enhanced language understanding and generation capabilities \cite{Brown2020-rh}. Besides the GPT series, companies like Google and Meta have initiated a steady release of large language models, ranging from tens to hundreds of billions, encompassing BERT, T5, RoBERTa, mT5, and other similar models \cite{Xue2020-pw,Najafi2022-kk}. Presently, notable open-source large models comprise Megatron, Turing-NLG, DALL-E \cite{lin2022large,Thirunavukarasu2023-vc,hadzic2023lateral}, along with domestic models like NeZHA, SuperCLUE, GLM-130B, ChatGLMM2 \cite{Wei2021-rv,Xu2023-la,Zeng2022-ki,10385748}.

\begin{figure*}[!t]%[H]
	%\centering
	%\includegraphics[width=0.7\linewidth]{Figure4.pdf}
	\centerline{\includegraphics[width=1.5\columnwidth]{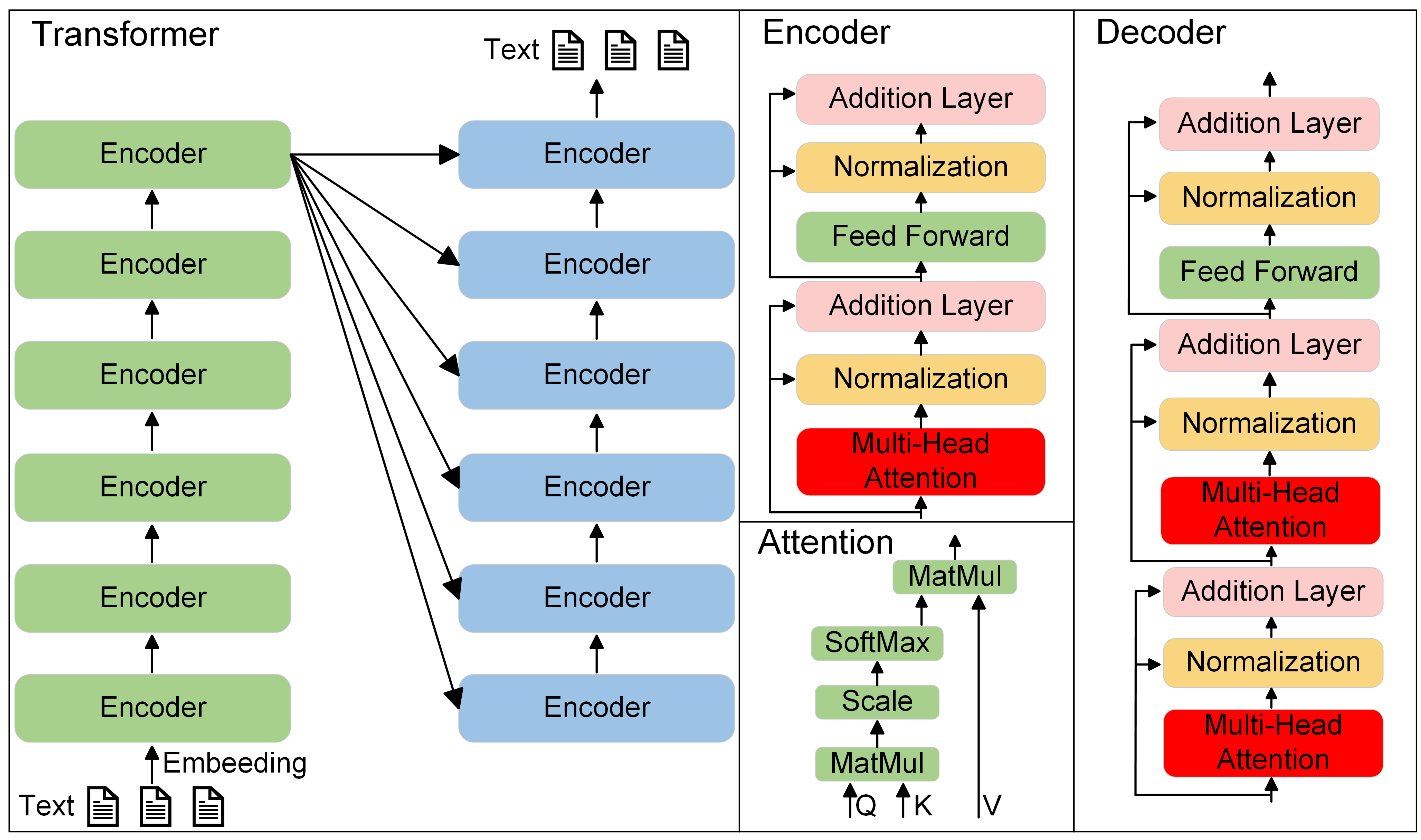}}
	\caption{The basic framework of transformer, including the structure of encoder and decoder and the technical details of the attention mechanism.}
	\label{fig:transformer}
	%\Description{Pipelines of robust watermarking, fragile watermarking, and our SepMark.}
\end{figure*}

Transformer is predominantly employed as the foundational framework in large models. The Transformer is a novel neural network architecture solely reliant on the attention mechanism, departing from the conventional structures of recurrent or convolutional neural networks. It builds upon preexisting sequence-to-sequence models, employing a blend of encoders and decoders. The transformer framework and its key technical details are illustrated in Figure~\ref{fig:transformer}. The transformer framework primarily consists of six encoder and decoder stacks. Initially, a substantial amount of text is embedded to transform high-dimensional textual information into a low-dimensional vector space, endeavoring to retain the semantic information-vocabulary relationship. The encoder primarily comprises a multi-head attention mechanism, a normalization layer \cite{xu2019understanding}, an addition layer, and a feedforward layer \cite{eldan2016power}. As depicted in Figure~\ref{fig:transformer}, the encoder efficiently extracts feature information from the input sequence via a multi-head attention mechanism, capturing crucial patterns and structures within the sequence \cite{tao2018get}. The multi-head attention mechanism enables the encoder to establish meaningful contextual connections across different positions, facilitating a deeper comprehension of the relationships between various segments within the input sequence \cite{naseem2020transformer}. Each encoder layer incorporates residual connections and normalization operations to mitigate gradient vanishing issues and expedite model training. The encoder's ultimate output is a latent representation containing the semantic information of the input sequence, which can be transmitted to the decoder or other tasks for additional processing. The decoder encompasses all encoder modules but varies in parameters. The decoder's self-attention mechanism models the relationships between different positions in the target sequence to enhance comprehension of its structural and semantic information. Additionally, the decoder introduces positional encoding to differentiate words or tokens at various positions within the target sequence. During sequence generation, the decoder generates each word or token in the target sequence iteratively, with each step depending on the preceding output and the hidden representation produced by the encoder. The pivotal component is the attention mechanism, employing three inputs: query (Q), key (K), and value (V), to compute the similarity between Q and V. Subsequently, the mechanism employs this similarity as a weighting factor for V, culminating in the weighted sum as output \cite{tao2018get}. This weighted sum denotes the attention degree that the model allocates to each position or feature in the input sequence, enabling automatic and selective emphasis on pertinent information.

The model framework can be categorized into five types: \ding{172} Solely comprising the transformer decoder component, typically employed for generative tasks like text generation and machine translation, represented by models like GPT and LLaMA \cite{zhang2023survey}. \ding{173} Solely encompassing the transformer encoder component, utilized for tasks such as text classification and sequence annotation, with BERT as a prominent representative. \ding{174} Encompassing both the encoder and decoder structures, constituting the complete transformer architecture, represented by models such as T5 and BERT, commonly applied in sequence-to-sequence tasks like machine translation and summarization. \ding{175} Incorporating the cross-transformer module, i.e., the cross-modal transformer structure, situated between the encoder and decoder to handle cross-modal data like text, images, and speech, with LXMERT being a notable large model example \cite{Hashemi2023-kh}. \ding{176} Visual transformer large model, dividing the image into serialized blocks and processing it using the transformer model, commonly utilized for generating video images, exemplified by Sora \cite{lu2023vdt}. These five types of large models are summarized in Table 1.

\begin{table*}[H]
	\caption{Summarize the representative large models among the five types of large models} 
	\label{tab:hyper}
	\renewcommand{\arraystretch}{1}
	\centering
	\resizebox{1\linewidth}{!}{
		\begin{tabular}{c|c}
			\hline
			\textbf{category} & \textbf{Models} \\
			\hline
			Decoder only           & {GPT1 \cite{radford2018improving}, GPT2 \cite{radford2019language},GPT3 \cite{brown2020language}, BART \cite{lewis2019bart}, LXM-R \cite{conneau2019unsupervised}, GPT-3.5 \cite{ouyang2022training}, LLaMA \cite{li2024llava}, Megatron-Turing-NLG \cite{smith2022using},T5 \cite{2020t5}}        \\
			Encoder only               & {BERT \cite{devlin2018bert}, RoBERTa \cite{zhuang2021robustly}, DistilBERT \cite{sanh2019distilbert}, ALBERT \cite{lan2019albert}, XLNet \cite{yang2019xlnet}, ERNIE\cite{sun2020ernie}, Electra \cite{clark2020electra}, CamemBERT \cite{martin2019camembert}, MT-DNN \cite{liu2019multi}}    \\
			Decoder and Encoder            & {MASS \cite{song2019mass}, Marian \cite{junczys2018marian}, M2M-100 \cite{fan2021beyond}, TAPAS \cite{herzig2020tapas}}\\
			Cross transformer        & {ViT-BERT\cite{li2021towards}, CLIP \cite{radford2021learning}, UNIMO \cite{li2020unimo}, MTViT \cite{yan2022multiview}, LXMERT \cite{tan2019lxmert}, VinVL \cite{zhang2021vinvl}}  \\
			Vision Transformer              & {VDM \cite{ho2022video}, Make-A-Video \cite{singer2022make}, ImageVideo \cite{saharia2022photorealistic}, Video LDM \cite{blattmann2023align}, Gen2 \cite{esser2023structure}, Emu Video \cite{girdhar2023emu}, Stable Video Difuddion \cite{blattmann2023stable}, Sora \cite{liu2024sora}}      \\
			
			\hline
	\end{tabular}}
\end{table*}

\subsection{Introduction to the principles of AI large models}
\begin{figure*}[!t]%[H]
	%\centering
	%\includegraphics[width=0.7\linewidth]{Figure4.pdf}
	\centerline{\includegraphics[width=2\columnwidth]{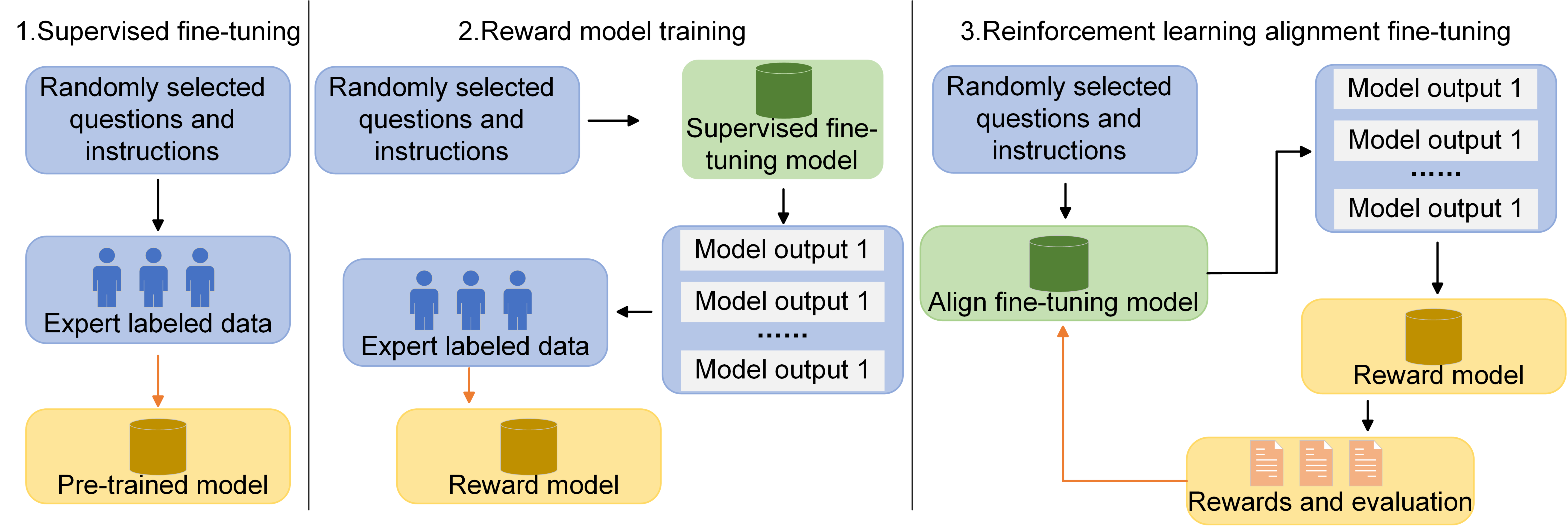}}
	\caption{Flow chart from large model training to fine-tuning}
	\label{fig:GPTflowchart}
	%\Description{Pipelines of robust watermarking, fragile watermarking, and our SepMark.}
\end{figure*}

To accomplish tasks in diverse complex scenarios such as natural language processing, computer vision, and speech recognition, large models necessitate pre-training on extensive unlabeled data followed by fine-tuning or end-to-end training on specific tasks \cite{Zhao2023-ku}. Large models aim to enhance the model's capacity to comprehend, adapt to human contexts, particularly healthcare settings, and address various challenges. The focal point lies in enabling the model to adeptly utilize the knowledge acquired during pre-training, fostering diverse capabilities to address a spectrum of challenges \cite{Zhao2024-ip}. The training process for a typical large model primarily comprises three steps: supervised training of the initial model, reward model training, and optimization through reinforcement learning \cite{Zhao2024-ip}. Figure.~\ref{fig:GPTflowchart} illustrates the flow chart depicting the process from large model training to fine-tuning. 

Supervised training of initial models stands as the predominant method for aligning large models with human preferences. This process leverages standard human-annotated datasets to train large models. Initially, a manually annotated dataset comprising input/output pairs needs to be gathered. Within this dataset, the input data constitutes the instructions or prompts provided to the model. Conversely, the output data comprises the responses anticipated from the model based on its exposure to extensive data, typically annotated by experts. Subsequently, the large model undergoes supervised fine-tuning using formatted data instructions. This method of supervised fine-tuning effectively enhances the large model's understanding of prior knowledge, enabling it to generate human-desired results efficiently.

In the training of large models, reward models typically involve designing a reward or loss function to steer model learning. Such a reward or loss function is devised to facilitate the model in optimizing a particular objective or task throughout training. Specifically, the objective of the reward model is to empower the large model to maximize or minimize the reward or loss function by adjusting its parameters. Consequently, throughout the training process, the large model updates parameters based on feedback from the reward or loss function, thereby progressively converging towards the optimal solution.

Reinforcement learning is employed to optimize the learning process, guided by the signal provided by the reward model \cite{Shakya2023-gz}. The optimization process involves the large language model's action domain, focusing on the prediction vocabulary, while the status pertains to the presently generated content. The feedback signal from the reward model is conveyed to the large model through an optimization algorithm, ultimately ensuring the alignment of the model's predictions with human expectations.

While large models excel in various tasks, their training typically demands substantial computational resources, including GPU and TPU utilization \cite{Wang2019-wv}. Due to their extensive parameters and intricate network structures, training large models is time-consuming, often spanning several days or even months. Of utmost concern is the tendency of large models to produce "hallucination" content in their outputs. This issue presents a challenge to the reliability of the model's practical applications. This "illusion" not only undermines user trust but also erodes confidence in the model. To mitigate hallucination occurrences, various approaches such as regularization, constraint augmentation, and multimodal training methods are employed to enhance the reliability of large model outputs.

\subsection{Multimodal AI large model}
\begin{figure*}[!t]%[H]
	%\centering
	%\includegraphics[width=0.7\linewidth]{Figure4.pdf}
	\centerline{\includegraphics[width=1.5\columnwidth]{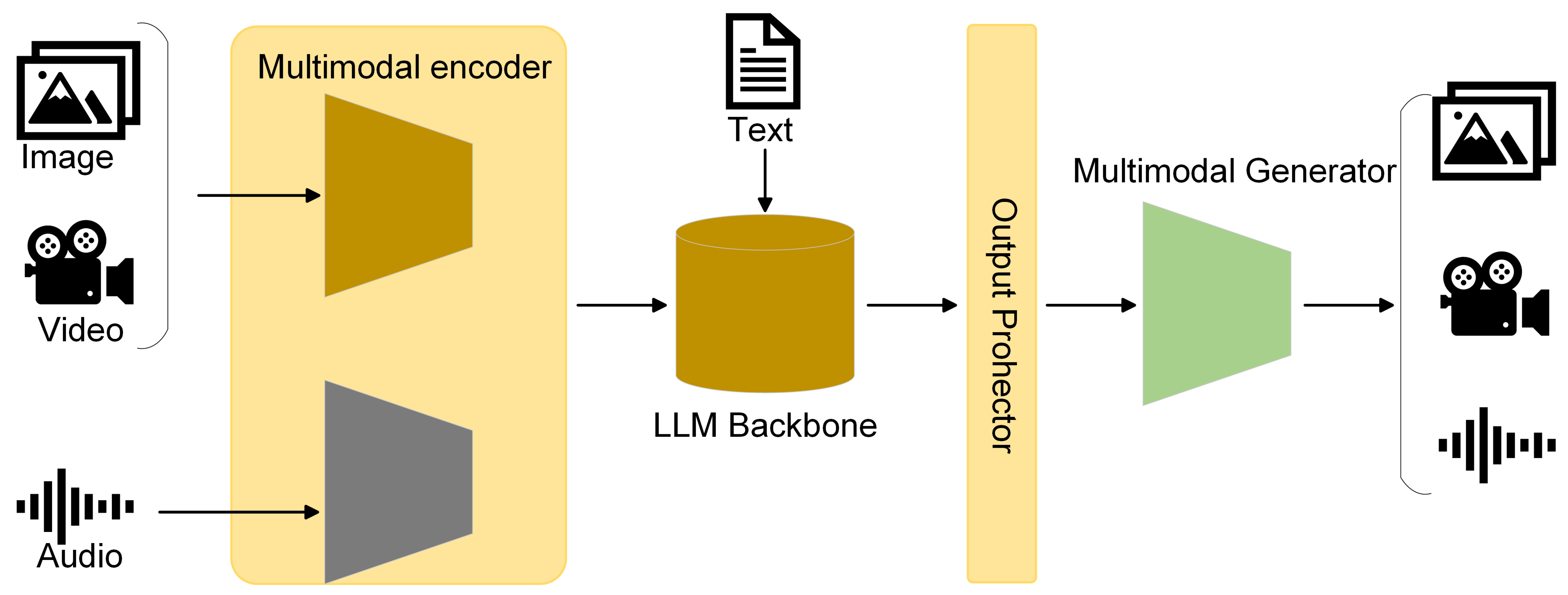}}
	\caption{The process of multimodal AI large model training and prediction \cite{Kruse2022-yo}.}
	\label{fig:MMLLM}
	%\Description{Pipelines of robust watermarking, fragile watermarking, and our SepMark.}
\end{figure*}

Multimodal large models offer numerous advantages compared to ChatGPT. They find extensive applications across various tasks and domains, encompassing comprehensive utilization of multi-source information, enriched semantic information, enhanced model robustness, and expanded applicability. It comprises three fundamental components: a visual encoder, a language model, and an adapter module. The primary role of the visual encoder is to process and comprehend input visual data, such as images. It extracts image features from pre-trained vision models, such as vision transformers or other convolutional neural network architectures. The language model serves as the core component of the multimodal large model, typically adopting transformer-based architectures such as BERT or the GPT series. Language models process textual inputs, facilitating comprehension and generation of natural language. The adapter module also plays a crucial role in the large multimodal model. It is responsible for bridging the gap between vision and language. The adapter module may take the form of a simple linear layer, a complex multi-layer perceptron, or a transformer, facilitating alignment between vision and text through the self-attention mechanism \cite{Kruse2022-yo,Pan2022-wv}. The process of training and predicting with multimodal AI large models is illustrated in Figure.~\ref{fig:MMLLM}.

During the initial training phase, multimodal large models must align text from diverse sources with visual data. This process involves integrating and fusing data from textual and visual modalities, aligning them within a unified representation space to enhance the model's comprehension of semantic relationships across modalities. In the subsequent training stage, instruction fine-tuning is employed to enhance the multimodal dialogue capabilities of the model. This involves fine-tuning the model using a dataset containing instructions, which may involve supervised fine-tuning or reinforcement learning based on human feedback, aiming to enhance the model's performance on a specific task \cite{wu2024fine}. These instruction datasets frequently comprise task-specific templates or prompts.

Multimodal (MM) LLMs, which are proficient in language modalities, face the primary challenge of effectively integrating information from diverse modalities to synthesize multimodal data for collaborative reasoning. The ultimate objective is to ensure that the outputs of MM LLMs align with human values \cite{Zhang2024-hh}.

\subsection{Video generation large model}
\begin{figure}[!t]%[H]
	%\centering
	%\includegraphics[width=0.7\linewidth]{Figure4.pdf}
	\centerline{\includegraphics[width=1\columnwidth]{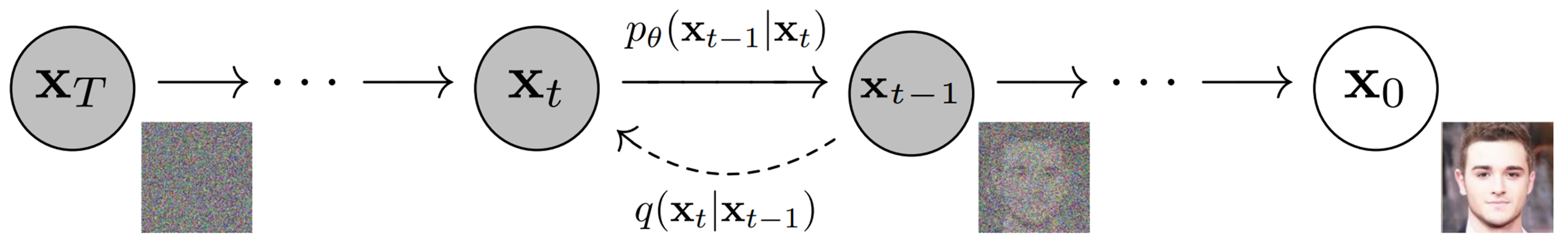}}
	\caption{The inverse process of diffusion learning for noisy images \cite{Liu2024-qv}.}
	\label{fig:diffusion}
	%\Description{Pipelines of robust watermarking, fragile watermarking, and our SepMark.}
\end{figure}

ChatGPT is a large-scale artificial intelligence language model. It utilizes an embedding layer to encode human language into its internal representation. Subsequently, it extracts rich knowledge and structures from vast datasets through the attention mechanism, synthesizing language output through weighted accumulation and association. The synthesized language output is decoded back into human-readable format. However, large-scale video generation models, exemplified by Sora, have sparked significant interest in the image domain. Sora is a diffusion transformer that introduces Gaussian noise continuously to corrupt the training data, subsequently learning to reconstruct the data by reversing this noise addition process \cite{Liu2024-qv}. Following training, the diffusion model can generate data by passing randomly sampled noise through a learned denoising process. The diffusion model is a latent variable that progressively introduces noise to the data to estimate an approximate posterior probability distribution \cite{croitoru2023diffusion}. The image undergoes a gradual transformation into pure Gaussian noise. The objective of training a diffusion model is to learn the reverse process. Traversing backward along this process chain enables the generation of new data, as illustrated in Figure.~\ref{fig:diffusion}.

From the perspective of information entropy, structured information exhibits low entropy \cite{yuan2021fuzzy}. Multiple rounds of Gaussian noise are added to elevate its information entropy, gradually concealing the original structural information. The pre-existing disordered unstructured segment exhibits high information entropy. Even without adding Gaussian noise, a small amount of it results in significant disorder.

A basic diffusion model, devoid of dimensionality reduction or compression throughout the process, exhibits a relatively high degree of reduction. The probability distribution in the learning process is parameterized as a latent variable. Its approximate distribution is obtained through training, and the distance between probability distributions is calculated using KL divergence \cite{Kim2021-zl}. The Diffusion transformer (DiT) incorporates transformers to perform multi-layer multi-head attention and normalization, thereby introducing dimensionality reduction and compression \cite{peebles2023scalable,pan2023ldcsf}. The process of negative information extraction in diffusion mode follows the same principle as the renormalization of LLM.

Google Lumiere also adopts a diffusion model and utilizes stacked normalization and attention layers, akin to Sora's DiT. However, it incorporates additional details such as duration, resolution, aspect ratio, etc. \cite{Bar-Tal2024-dq}. These aspects are handled differently. Success or failure hinges on the specifics. OpenAI stated that Sora departed from the common practice of resizing, cropping, or trimming videos to standard sizes, as observed in other Wensheng videos. Instead, it trained video generation with variable duration, original resolution, and aspect ratio to attain significant benefits, including flexible sampling \cite{noauthor_undated-wv}.

\section{Applications of AI large model in radiology}
\subsection{Radiology education}
Nations worldwide prioritize the research and implementation of expansive educational models. In May 2023, the U.S. Department of Education's Office of Educational Technology issued a comprehensive policy report titled "Artificial Intelligence and the Future of Teaching and Learning," aimed at fostering the integration of AI large models into education. On March 29, 2023, the British Parliament reviewed the policy document "Regulatory Approach to Supporting AI Innovation," stressing the role of AI in accelerating the development of diverse industries and overcoming energy efficiency challenges. On September 7, 2023, UNESCO issued its inaugural global guidelines advocating for the integration of generative AI in education, urging nations to enact relevant initiatives. In September 2023, Hong Kong, China, introduced an artificial intelligence curriculum tailored for junior high school students, mandating that public schools provide 10 to 14 hours of AI instruction, covering subjects such as ChatGPT, AI ethics, and the societal implications of AI.

Informatization is increasingly driving innovation in medical education. Currently, information delivery is transitioning from traditional "paper-based" to "electronic media," thereby necessitating changes in the conventional paper-based educational processes. Informatization is reshaping learning, classroom, and assessment models, and even leading to the emergence of "Process Reengineering". AI large models have demonstrated their potential in medical professional education across various disciplines including mathematics, engineering, and art. In certain exams, ChatGPT offers interpretable responses to specialized medical queries, showcasing narrative coherence. Hilal et al. compared dental students' performance with that of ChatGPT in oral and maxillofacial radiology, revealing ChatGPT's limited proficiency and applicability in actual examinations within this field \cite{Ozturk2023-if}. Namkee et al. assessed ChatGPT's ability to comprehend complex surgical clinical data and its potential implications for surgical education and training \cite{Oh2023-pb}. They observed that GPT4 exhibited remarkable proficiency in comprehending intricate surgical clinical data, achieving an accuracy rate of 76.4\% in the Korean General Surgery Board Examination \cite{Oh2023-pb}. Ian J et al. conducted two experiments to evaluate three different chatbots' responses to ten questions related to CT, MRI, and bone biopsy. Two independent reviewers assessed the accuracy and completeness of the chatbot responses \cite{Kuckelman2023-lv}. The experiments revealed that the Bing large model yielded precise responses, devoid of inaccuracies or potential user confusion \cite{Kuckelman2023-lv}. GPT4-turbo's clinical accuracy on 300 exam questions across four main domains (clinical, biology, physics, and statistics) matches that of high-level students and surpasses that of low-level students \cite{Thaker2024-yi}. G et al. evaluated the model's ability to accurately answer questions in five knowledge areas on 1064 alternative questions simulating a health physics certification exam \cite{Roemer2024-sz}. Analysis shows that although the overall accuracy of GPT4 is higher, the answers in GPT3.5 format are more correct \cite{Roemer2024-sz}.

\begin{figure*}[!t]%[H]
	%\centering
	%\includegraphics[width=0.7\linewidth]{Figure4.pdf}
	\centerline{\includegraphics[width=2.1\columnwidth]{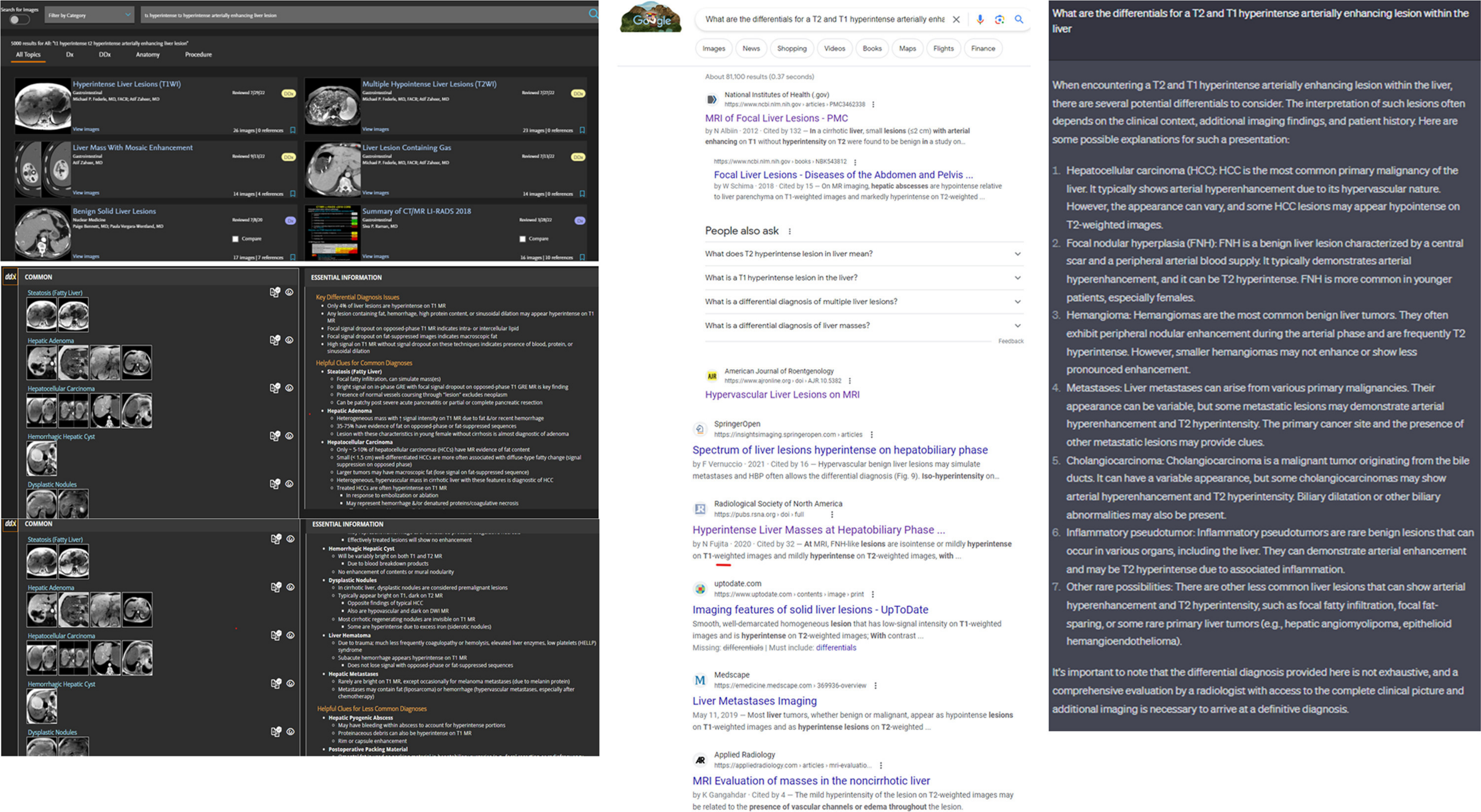}}
	\caption{Screen shots depict a typical query conducted on various search engines, including StatDx (a widely-used radiology search engine for analyzing complex or unusual cases) and Google, contrasted with a standard ChatGPT response to a query regarding the "differential diagnosis of a T2 and T1 hyperintense arterially enhancing lesion of the liver". The ChatGPT response is centralized, concise, and suitable for enhancing the learning experience of radiology trainees in the reading room. For the complete conversation, refer to the following link: \url{https://chat.openai.com/share/88f09912-1cef-4cf0-a488-bf97086f9125} \cite{tippareddy2023radiology}.}
	\label{fig:education}
	%\Description{Pipelines of robust watermarking, fragile watermarking, and our SepMark.}
\end{figure*}

Following extensive training with specialized medical knowledge, AI large models can establish an interactive learning environment where young radiologists can pose questions on demand, as depicted in Figure.~\ref{fig:education}. Differential diagnosis and sign analysis will equip young radiologists with valuable insights for their daily clinical practice. It serves as a platform for continuous learning. In clinical practice, AI large models can analyze imaging reports in real time, assisting radiologists in employing suitable descriptive language and enhancing the quality management of report documents. AI large models can simplify explanations of complex imaging concepts and findings, facilitating trainees' comprehension and practical application. AI large models hold the potential to shape future curriculum design, educational planning, and teaching methodologies for imaging educators. For instance, ChatGPT can assist educators in drafting lesson plans, generating interactive Q\&A sessions, clinical samples, and more. The integration of AI large models into medical education holds significant promise for enhancing students' learning experiences and fostering a more interactive and engaging educational environment.

\subsection{Radiology report generation}
Radiology images encompass abundant pathological information, including X-rays, CT scans, MRIs, and more. Doctors cannot deliver precise diagnostic reports at all times. Medical AI large models can assimilate vast amounts of imaging data to ensure prompt processing of patients' diagnostic needs, alleviate the diagnostic burden on radiologists, and enhance diagnostic efficiency and accuracy. Concurrently, through continuous enhancement and refinement of functionalities tailored to diverse regions, linguistic features, diagnostic practices, etc., AI large models can cultivate digital radiologists capable of adjusting to varied diagnostic and treatment idiosyncrasies and regional nuances, particularly in areas with less-than-ideal medical infrastructure. Furthermore, they can access expert-level imaging diagnoses from tertiary hospitals, thereby elevating the standard of medical services.

\begin{figure}[!t]%[H]
	%\centering
	%\includegraphics[width=0.7\linewidth]{Figure4.pdf}
	\centerline{\includegraphics[width=1\columnwidth]{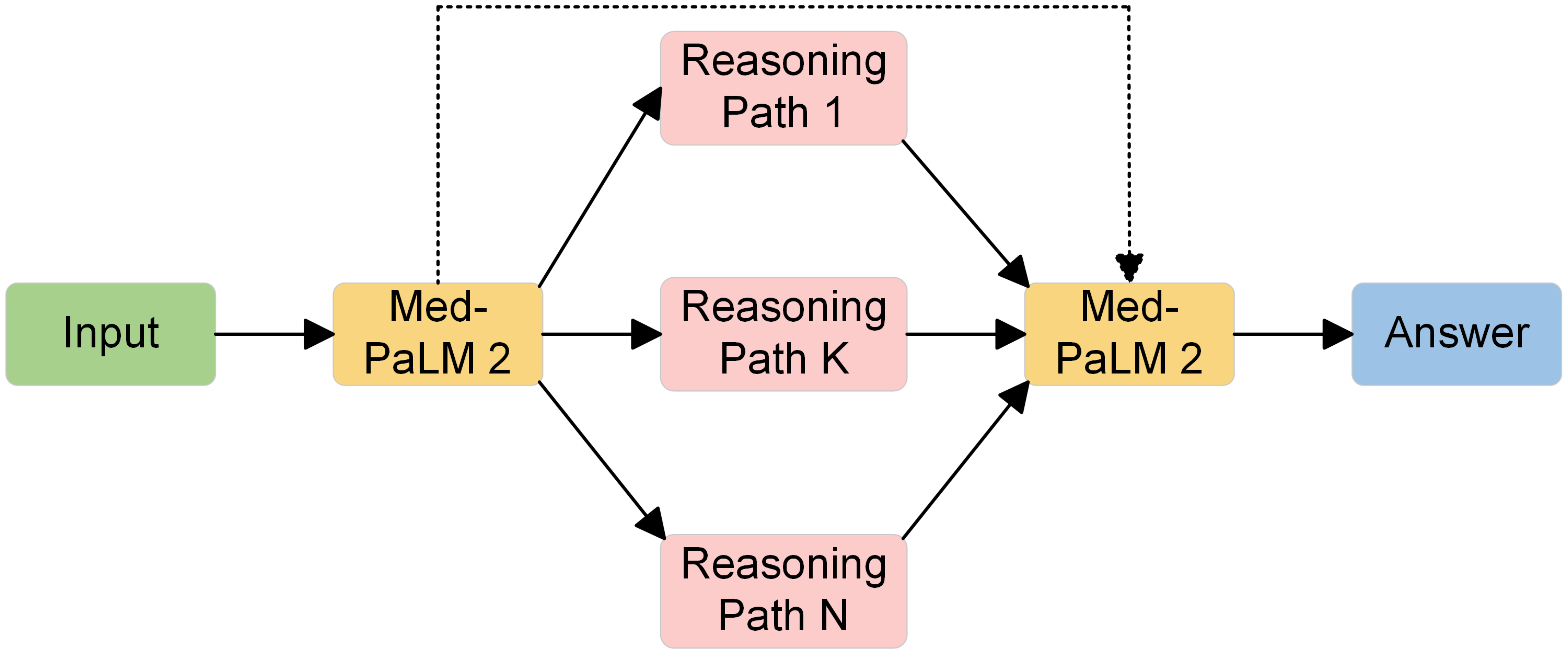}}
	\caption{Use Med-PaLM 2 to diagnose and generate standard answers \cite{Singhal2023-qh}.}
	\label{fig:Med-Palm2}
	%\Description{Pipelines of robust watermarking, fragile watermarking, and our SepMark.}
\end{figure}

Given the ongoing advancements in AI medical diagnosis, single-disease diagnosis fails to adequately address the needs of both physicians and patients. Multi-disease and multi-task AI models are currently under development. AI large models can effectively and quantitatively analyze images and transcribe imaging reports into medical records. However, these descriptions often remain superficial, fail to lead to diagnostic conclusions, and sometimes result in incorrect diagnoses. In 2023, Google proposed Med-PaLM2 with the aim of furnishing high-quality responses to medical inquiries, as depicted in Figure.~\ref{fig:Med-Palm2} \cite{Singhal2023-qh}. Med-PaLM can additionally produce accurate, informative, detailed responses to consumer health queries, drawing on input from physicians and user panels \cite{Singhal2023-tg}. On April 12, 2023, Nature published an article introducing a novel paradigm for medical artificial intelligence known as general medical artificial intelligence (GMAI) \cite{Moor2023-be}. The emergence of GMAI has enabled researchers in the medical field to recognize the significant potential of AI in transforming the entire healthcare system. In the future, following extensive fine-tuning with professional imaging knowledge, AI large models are anticipated to generate provisional diagnostic conclusions solely based on imaging report descriptions and potential connections between multiple lesions. GatorTron is an electronic health record (EHR) big data model developed at the University of Florida, developed from scratch as an LLM (no other pre-trained models based on it), improved using 8.9 billion parameters and more than 90 billion words of text from electronic health records five clinical natural language processing tasks, including medical question answering and medical relationship extraction \cite{Yang2022-cu}.

\begin{figure*}[!t]%[H]
	%\centering
	%\includegraphics[width=0.7\linewidth]{Figure4.pdf}
	\centerline{\includegraphics[width=2.1\columnwidth]{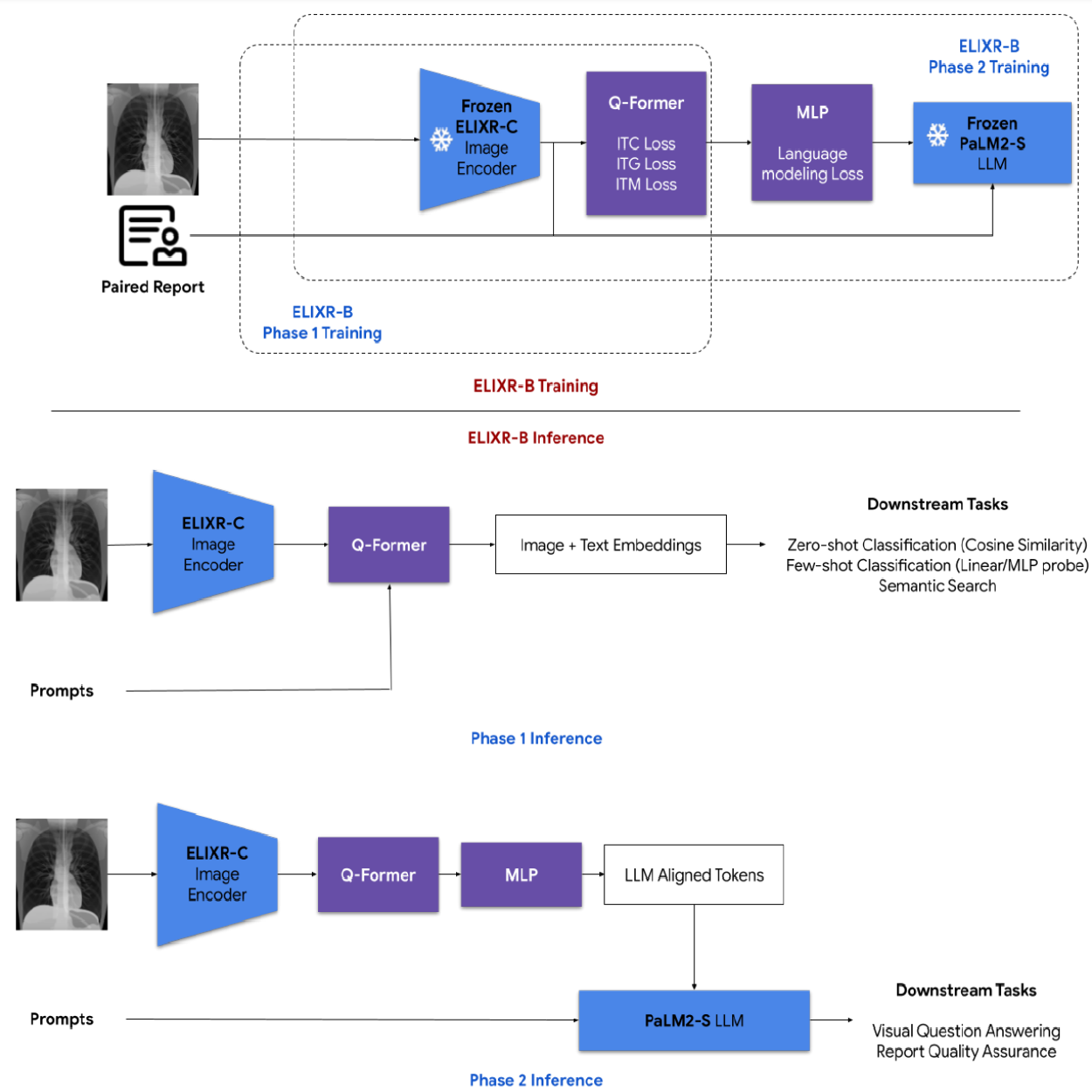}}
	\caption{ELIXR Architecture. (a) Training and inference of ELIXR-C. (b) Training and inference of ELIXR-B. ELIXR-B is trained in two phases. In the first phase, the model utilizes three learning objectives (image-text contrastive learning (ITC), image-grounded text generation (ITG), image-text matching (ITM) losses) to bootstrap vision-language representation in the Q-Former, learning from embeddings obtained from a frozen image encoder. In the second phase, the model bootstraps vision-to-language generation from a frozen large language model. The purple text boxes represent the learned (unfrozen) components during the training step. Details of the VQA inference are further elaborated in the relevant section \cite{Xu2023-cp}.}
	\label{fig:EXLXR}
	%\Description{Pipelines of robust watermarking, fragile watermarking, and our SepMark.}
\end{figure*}

Using large models to generate radiology reports is a current research trend. In their study on generating chest radiology reports, Wang et al. introduced R2GenGPT, which employs an efficient visual alignment module to align visual features with the word embedding space of LLM. This allows the previously static LLM to seamlessly integrate and process image information, resulting in improved performance on two benchmark datasets \cite{Wang2023-px}. To safeguard user privacy, Pritam et al. introduced Vicuna for labeling radiology reports. They achieved improved results on chest X-ray radiography reports in MIMIC-CXR and National Institutes of Health (NIH) datasets \cite{Mukherjee2023-aq}. ELIXR employs a language-aligned image encoder combined with the fixed LLM PaLM 2 to perform various chest X-ray tasks, achieving an average AUC of 0.893 for zero-sample chest X-ray (CXR) classification and 0.898 for data-efficient CXR classification \cite{Xu2023-cp}. The training and inference architecture of ELlXR is illustrated in Figure.~\ref{fig:EXLXR} \cite{Xu2023-cp}. Chantal et al. proposed RaDialog, which effectively integrates visual image features and structured pathology results with a large language model (LLM). They adjusted it to professional domains through efficient fine-tuning of parameters \cite{Pellegrini2023-vx}. Xu et al. utilized LLMs to enhance semantic analysis and develop similarity metrics for text. They addressed the limitations of traditional unsupervised NLP metrics (e.g., ROUGE and BLEU) and demonstrated the potential of using LLMs for semantic analysis of textual data, leveraging semiquantitative inference results from highly specialized domains \cite{Xu2024-ih}. Stephanie et al. introduced the MAIRA-1 model, which employs a CXR-specific image encoder combined with a fine-tuned large language model based on Vicuna-7B and text-based data augmentation to produce reports of state-of-the-art quality \cite{Hyland2023-yc}. Specifically, MAIRA-1 significantly enhanced the radiologist-aligned RadCliQ metric and all considered lexical metrics \cite{Hyland2023-yc}. Jawook et al. developed a BERT-based tagger called CheX-GPT, which operates faster and more efficiently than its GPT counterpart \cite{Gu2024-fb}. Additionally, CheX-GPT surpasses existing models in labeling accuracy on the expert annotation test set MIMIC-500 and exhibits superior efficiency, flexibility, and scalability \cite{Gu2024-fb}. Li et al. developed an open-source multimodal large language model (CXR-LLAVA) for interpreting chest X-ray images (CXR), leveraging recent advances in large language models (LLM) to potentially replicate human radiologists’ Image interpretation skills materials and methods, achieving better results than GPT-4-vision and Gemini-Pro-Vision on two training data sets and one testing data set \cite{Lee2023-yh}. Ali H et al. proposed Domain Adaptive Language Modeling (RadLing) to extract Common Data Elements (CDEs) from chest radiology reports, which comprehensively outperforms the performance of GPT4 in terms of accuracy, recall and is easy to deploy locally with low running costs \cite{Dhanaliwala2023-in}.

The improvement efforts based on large models persist beyond this point. Zhu et al. employed contextual instructional learning (ICIL) and chain of thought (CoT) reasoning methods to integrate the expertise of professional radiologists with large language models (LLM). Evaluating LLMs and aligning them with radiologist standards has the potential to enhance the quality assessment of AI-driven medical reporting \cite{Zhu2024-tj}. A. Infante et al. investigated ChatGPT, Bard, and Perplexity for extracting emergency data compared to human experts. They found that LLM outperformed in the medical field \cite{Infante2023-cc}. Katharina et al. surveyed 15 radiologists, asking them to assess the quality of simplified radiology reports based on factual correctness, completeness, and potential patient risk. They found that employing LLMs like ChatGPT enhanced radiology and other patient-centered care in medicine \cite{Jeblick2023-cl}. Li et al. utilized OpenAI ChatGPT to improve patients' comprehension of diagnostic reports. They discovered that the model exhibited average reading levels comparable to those of readers in a stratified sample of radiology reports spanning X-rays, ultrasound, CT, and MRI \cite{Li2023-wd}. In collaboration with the Second Xiangya Hospital, Zhong et al. introduced ChatRadio-Valuer, an LLM-based system for the automatic generation of customized models for radiology reports. These models can learn generalizable representations and serve as a foundation for model adaptation in complex analytical scenarios \cite{Zhong2023-mm}. ChatRadio-Valuer consistently surpasses state-of-the-art models in disease diagnosis in radiology reports. This effectively enhances model generalization performance and reduces the workload of expert annotation, thereby promoting the application of clinical AI in radiology reports \cite{Zhong2023-mm}. Yan et al. integrated RadGraph, a graphical representation of reports, with the Large Language Model (LLM) to extract content from images. They subsequently verbalized the extracted content into reports tailored to the style of specific radiologists \cite{Yan2023-sd}. Lu et al. introduced a simple yet effective two-stage fine-tuning protocol, based on the OpenLLaMA-7B framework, to spatially align visual features with LLM's text embeddings as soft visual cues \cite{Lu2023-vy}. Additionally, a detailed analysis of soft visual cues and attention mechanisms was conducted, inspiring future research directions \cite{Lu2023-vy}. Work by Takeshi et al. found no significant differences (p > 0.05) in qualitative scores between radiologists and GPT-3.5 or GPT-4 in terms of syntax and readability, image discovery, and overall quality \cite{Nakaura2023-yh}. However, the GPT series had significantly lower qualitative scores than radiologists in terms of impression and differential diagnosis scores (p < 0.05) \cite{Nakaura2023-yh}.

Numerous studies are also conducted on generating various radiology reports. GPT4 can aid orthopedic surgeons in classifying fracture morphology, achieving a consistent accuracy of 71\% \cite{Russe2023-bn}. While GPT4 may not match human accuracy, its speed surpasses humans significantly. Furthermore, providing domain-specific knowledge can significantly enhance GPT's performance and consistency. Eric M et al. utilized ChatGPT to summarize five complete MRI reports of prostate cancer patients conducted at a single institution from 2021 to 2022. They generated 15 summary reports tailored to the reading level of the patients \cite{Chung2023-vx}. Jaeyoung et al. employed a large language model (LLM) to integrate multiple image analysis tools into the breast reporting process using LangChain \cite{Huh2023-to}. Through the combination of specific tools and LangChain text generation, their approach accurately extracts relevant features from ultrasound images, interprets them in clinical settings, and generates comprehensive and standardized reports \cite{Huh2023-to}. In an analysis of 99 radiology reports, ChatGPT achieved a final diagnostic accuracy of 75\% (95\% CI: 66-83\%), compared to the 64\%-82\% accuracy range of radiologists \cite{Mitsuyama2023-vt}. ChatGPT demonstrates strong diagnostic capabilities and is comparable to neuroradiologists in distinguishing brain tumors in MRI reports. It can serve as a secondary opinion for neuroradiologists' final diagnoses and a guiding tool for general radiologists and residents, particularly in understanding diagnostic clues and managing complex cases \cite{Mitsuyama2023-vt}. In the future, numerous large AI models will be developed and utilized in medical imaging diagnostic reports.

\subsection{Applications of unimodal radiology}
Medical imaging encompasses various tasks, including segmentation, classification, and detection, assisting doctors in efficiently identifying and diagnosing patients' conditions. The segmentation task precisely delineates anatomical structures and lesion areas in medical images. Segmentation results offer doctors detailed anatomical information, aiding in lesion localization, size measurement, and surgical planning. The classification task entails identifying lesion type, disease stage, or tissue type within images through feature learning and classification in medical imaging. Classification results aid doctors in making initial diagnoses, distinguishing between diseases, and selecting appropriate treatment options. The detection task involves automatically identifying specific lesions or abnormal areas in medical images, such as tumors, stones, and hemangiomas. Test results assist doctors in identifying potential lesions and facilitating early diagnosis and treatment.

\begin{figure*}[!t]%[H]
	%\centering
	%\includegraphics[width=0.7\linewidth]{Figure4.pdf}
	\centerline{\includegraphics[width=2.1\columnwidth]{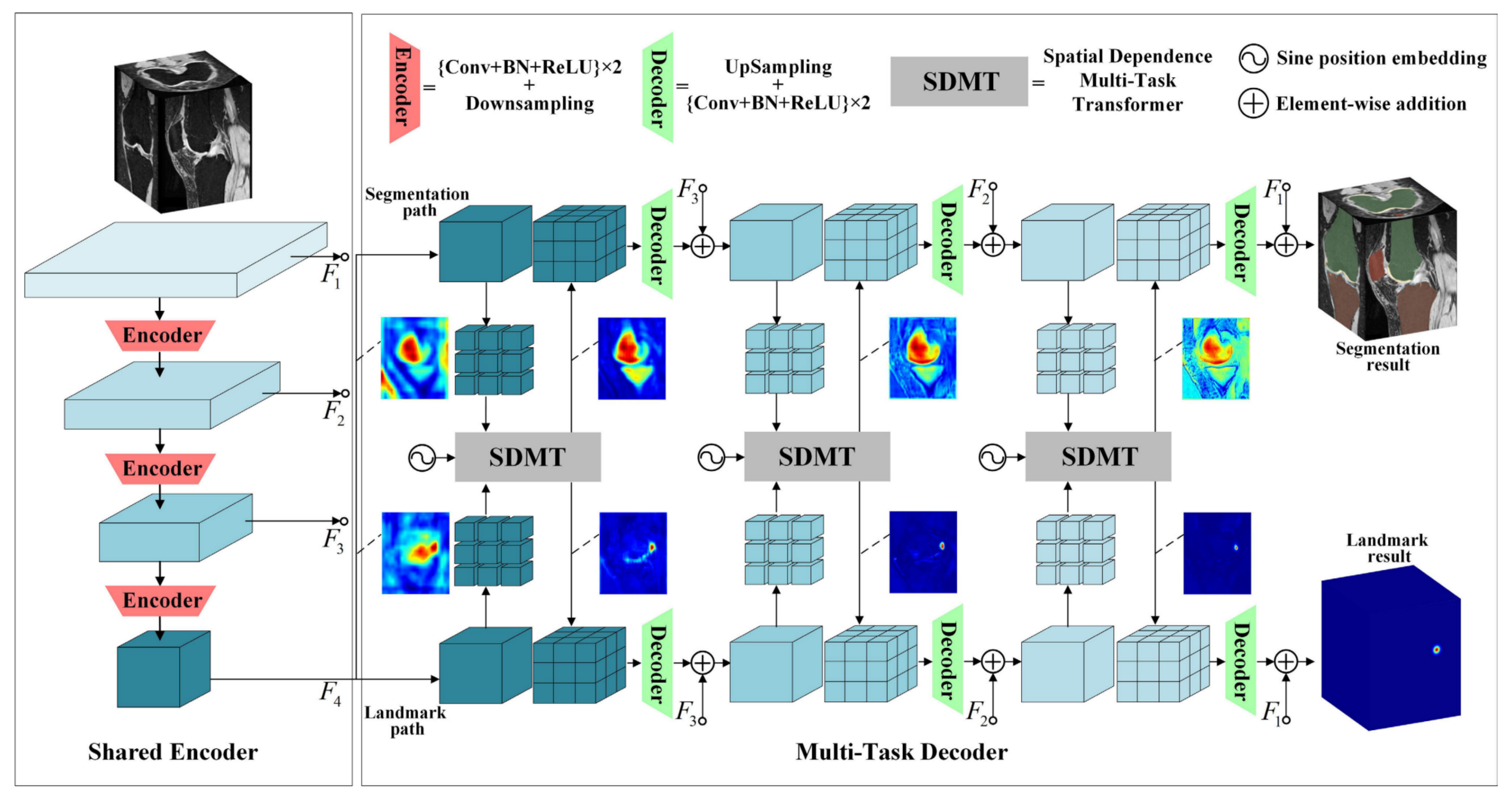}}
	\caption{Network Structure Overview: The proposed method begins by employing a shared encoder to extract multi-scale features {F1, F2, F3, F4}. Subsequently, the decoder stage is divided into two paths: segmentation and landmark paths. The Spatial Dependency Modeling Transformer (SDMT) is positioned between these paths to capture spatial dependencies across both tasks and rectify features accordingly. Finally, each task path independently decodes its respective task to produce outputs \cite{li2023sdmt}.}
	\label{fig:SDMT}
	%\Description{Pipelines of robust watermarking, fragile watermarking, and our SepMark.}
\end{figure*}

Within the segmentation task, the end-to-end ProstAttention-Net conducts comprehensive multi-class segmentation of prostate and cancer lesions based on Gleason score, demonstrating robust generalization \cite{duran2022prostattention}. A 4-dimensional (4D) deep learning model, employing 3D convolution and convolutional long short-term memory (C-LSTM), harnesses 4D data extracted from dynamic contrast-enhanced (DCE) magnetic resonance imaging (MRI) to facilitate liver tumor segmentation \cite{zheng2022automatic,pan2023cvfc}. As depicted in Figure.~\ref{fig:SDMT}, a spatially dependent multi-task transformer (SDMT) network is employed for 3D knee MRI segmentation and landmark localization \cite{li2023sdmt}. SDMT incorporates spatial coding into features and devises a task-mixed multi-head attention mechanism, wherein attention heads are categorized into inter-task and intra-task attention heads \cite{li2023sdmt}. These attention heads manage spatial inter-dependencies between tasks and intra-task correlations, respectively \cite{li2023sdmt}. Edge U-Net, inspired by the U-Net architecture, is an encoder-decoder structure that enhances tumor localization by fusing boundary-related MRI data with primary brain MRI data \cite{allah2023edge}. SynthSeg+ is an AI segmentation suite that facilitates robust analysis of heterogeneous clinical datasets \cite{billot2023robust}. Apart from whole-brain segmentation, SynthSeg+ conducts cortical segmentation, estimates intracranial volume, and automatically detects mis-segmentations (primarily resulting from very low-quality scans) \cite{billot2023robust}.

Deep learning is commonly employed in classification tasks to categorize radiological images, yielding enhanced diagnostic outcomes. A CNN-based model attained an AUC of 0.9824 ± 0.0043, with accuracy, sensitivity, and specificity values of 94.64 ± 0.45\%, 96.50 ± 0.36\%, and 92.86 ± 0.48\%, respectively, in distinguishing between normal and abnormal chest radiographs \cite{tang2020automated}. An image-based model, constructed on the EfficientNet-B0 architecture, is coupled with a logistic regression model trained on patient demographics and lesion location to differentiate between benign and malignant lesions, yielding superior outcomes \cite{eweje2021deep}. Wang et al. introduced the triple attention network ($A^{3}$Net) for chest X-ray diagnosis of chest diseases, incorporating an attention mechanism. Utilizing pre-trained DenseNet-121 for feature extraction, they integrated three attention modules—channel, elemental, and scale approach—into a unified framework, achieving the highest average AUC of 0.826 per class across 14 chest diseases \cite{wang2021triple}. MBTFCNy, comprising feature extraction (FE), residual strip pooling attention (RSPA), atrous space pyramid pooling (ASPP), and classification modules, is effective for brain tumor MRI classification \cite{shahin2023mbtfcn}. Lastly, a customized MobileNetV2 achieves high-precision multi-class classification of lung diseases from chest X-ray images, surpassing models like InceptionV3, AlexNet, DenseNet121, VGG19, and MobileNetV2 \cite{shamrat2023high}.

Furthermore, numerous models are capable of concurrently performing multiple tasks. Compared to two radiology residents and two radiologists specialized in musculoskeletal training, a multi-task deep learning model demonstrated higher accuracy in diagnosing benign or malignant bone tumors across all patients \cite{von2021multitask}. A deep convolutional neural network (CNN)-based architecture was utilized for automatic classification of brain images into four classes, alongside a U-Net-based segmentation model for brain tumor MRI classification and segmentation. These models were evaluated on six benchmark datasets, with the segmentation model achieving superior performance in training \cite{akter2024robust}. Cerberus, a multi-task learning method, is employed for segmenting and classifying nuclei, glands, lumens, and tissue regions utilizing data from diverse independent sources \cite{graham2023one}. ResGANet employs ResNet and its variants as the backbone network, utilizing modular group attention blocks to capture feature dependencies in medical images across two independent dimensions (channel and space) \cite{cheng2022resganet}. The encoder is grounded on semi-supervised 3D depthwise convolution Inception, complemented by 3D squeeze, excitation blocks, and depthwise convolutions. Additionally, a residual learning-based decoder is employed for tumor cell segmentation, facilitating adaptive recalibration of channel features via explicit interdependence modeling and amalgamating coarse and intricate features, thus enabling precise tumor segmentation and survival prediction \cite{qayyum2023semi}.

The Segment Anything Model (SAM) has gained popularity, making it increasingly prevalent in professional settings. These models can autonomously process data following their acquisition of medical imaging knowledge \cite{Kirillov2023-hy}. Label-Studio SAMed, a solution for medical image segmentation, is built upon SAM technology \cite{Zhang2023-ej}. It customizes SAM for medical image segmentation by employing a fine-tuning strategy with low-rank benchmarks. Furthermore, two tracking models exist: the Track Anything Model (TAM) and SAM-Track \cite{wang2021unidentified,cheng2022pointly}. The expansion of these models into the medical domain can facilitate real-time processing of medical imaging data by physicians.

\subsection{Applications of multimodal radiology}
Medical imaging data offer abundant information encompassing various biological characteristics and tissue structures. Analyzing multimodal imaging data enables a more comprehensive understanding of the patient's condition and physiological status \cite{song2022multimodal}. Different types of medical imaging data are complementary, mutually reinforcing, and capable of confirming one another. Multimodal medical image analysis can optimally leverage the strengths of diverse image types to enhance the accuracy of disease diagnosis by physicians. For example, in tumor diagnosis, integrating CT and MRI images can yield comprehensive information on tumor characteristics, aiding in the determination of tumor type, size, and location. Multimodal medical image analysis offers substantial support for personalized medicine. Comprehensive analysis of patients' multimodal imaging data and other clinical information enables the development of personalized diagnosis and treatment plans, thereby enhancing treatment outcomes and quality of life. 

Various multimodal radiology applications have been extensively employed in tumor diagnosis. Zhang et al. proposed a novel strategy called mutual learning (ML) for effective and robust segmentation of multimodal liver tumors \cite{zhang2021modality}. In contrast to existing multimodal approaches that fuse information from various modalities using a single model, machine learning enables ensembles of modality-specific models to collaboratively refine features and commonalities between high-level representations of different modalities by learning from and teaching each other \cite{zhang2021modality}. Huang et al. introduced a novel framework called AW3M, which collectively employs four types of ultrasonography (i.e., B-mode, Doppler, shear wave elastography, and strain elastography) to assist in breast cancer diagnosis \cite{huang2021aw3m}. The effectiveness of the AW3M framework was also validated on multiple multimodal datasets \cite{huang2021aw3m}. Fu et al. presented a deep-learning-based framework for multimodal PET-CT segmentation featuring a multimodal spatial attention module (MSAM) \cite{fu2021multimodal}. MSAM automatically learns to highlight tumor-related spatial regions and suppress standard regions with physiologically high uptake based on PET input \cite{fu2021multimodal}. The resulting spatial attention maps are then utilized to guide a convolutional neural network (CNN) backbone in segmenting regions with a high likelihood of tumors from CT images \cite{fu2021multimodal}. Zhang et al. introduced a semi-supervised contrast mutual learning (Semi-CML) segmentation framework, in which a novel area similarity contrast (ASC) loss leverages cross-modality information between modalities to ensure consistency in contrast mutual learning \cite{zhang2023multi}. The results demonstrate that Semi-CML with PReL significantly outperforms state-of-the-art semi-supervised segmentation methods. It achieves performance similar to, and sometimes even better than, fully supervised segmentation methods with 100\% labeled data while reducing the data annotation cost by 90\% \cite{zhang2023multi}. Secondly, LViT (Language meet Vision Transformer), a novel text-enhanced medical image segmentation model, integrates medical text annotations to address the quality deficiencies of image data. It delivers excellent segmentation performance across all three multimodal medical segmentation datasets (image + text), which include X-ray and CT images \cite{li2023lvit}. Lastly, ResViT, a novel generative adversarial method for medical image synthesis, leverages the context sensitivity of the visual transformer, the accuracy of the convolution operator, and the fidelity of adversarial learning to synthesize missing sequences in multi-contrast MRIs and CT images from MRIs \cite{dalmaz2022resvit}. Experimental results demonstrate that ResViT surpasses CNN and transformer-based methods in both qualitative observations and quantitative metrics \cite{dalmaz2022resvit}.

Building a robust and comprehensive foundational model in the medical domain can offer smarter and more efficient solutions for clinical tasks, enhance the medical experience for both healthcare professionals and patients, and usher in a new era of technological innovation \cite{pan2024selector}. The effectiveness of multi-modal large models in general fields is relatively limited in radiology \cite{guo2019deep}. VisualGLM-6B is an open-source multi-modal conversational language model that supports image, Chinese, and English inputs \cite{du2021glm}. Based on ChatGLM-6B, this model boasts 6.2 billion parameters. It utilizes high-quality image-text pairs for training and fine-tuning on extensive visual Q\&A datasets to produce human-preference-consistent responses \cite{du2021glm}. Visual Med-Alpaca, based on the LLaMa-7B architecture, is trained using GPT-3.5-Turbo and a human expert-curated instruction set. Equipped with plug-and-play vision modules and extensive instruction tuning, it is capable of diverse tasks, ranging from interpreting radiology images to addressing complex clinical queries \cite{Reference}. XrayGLM is fine-tuned and trained based on VisualGLM-6B. With the assistance of ChatGPT, a pair of X-ray image-diagnosis and treatment reports is constructed to support Chinese training \cite{wang2023XrayGLM}. LLaVA-Med is a large-scale multimodal model relying on a comprehensive biomedical figure caption dataset. With excellent multimodal dialogue capabilities, it can follow open instructions to assist in querying information about biomedical images \cite{li2024llava}. Shanghai Artificial Intelligence Laboratory (Shanghai AI Laboratory) leads the initiative. It collaborates with top scientific research institutions, universities, and hospitals globally to jointly launch the world's first medical multimodal foundational model group, "OpenMEDLab" \cite{Wang2024-ec}. Among them, RadFM can support three-dimensional data, multi-image input, and interleaved data formats, significantly enhancing the clinical application of fundamental medical models \cite{Zhang2023-dp}. Peking University proposed Qilin-Med-VL, a large-scale visual language model, to integrate text and visual data analysis. It enhances the model's capability to generate headlines and answer complex medical queries \cite{Ye2023-fq,Liu2023-tw}. Additionally, multimodal medical imaging datasets will further advance research progress in multimodal medical large models.

\begin{figure}[!t]%[H]
	%\centering
	%\includegraphics[width=0.7\linewidth]{Figure4.pdf}
	\centerline{\includegraphics[width=1\columnwidth]{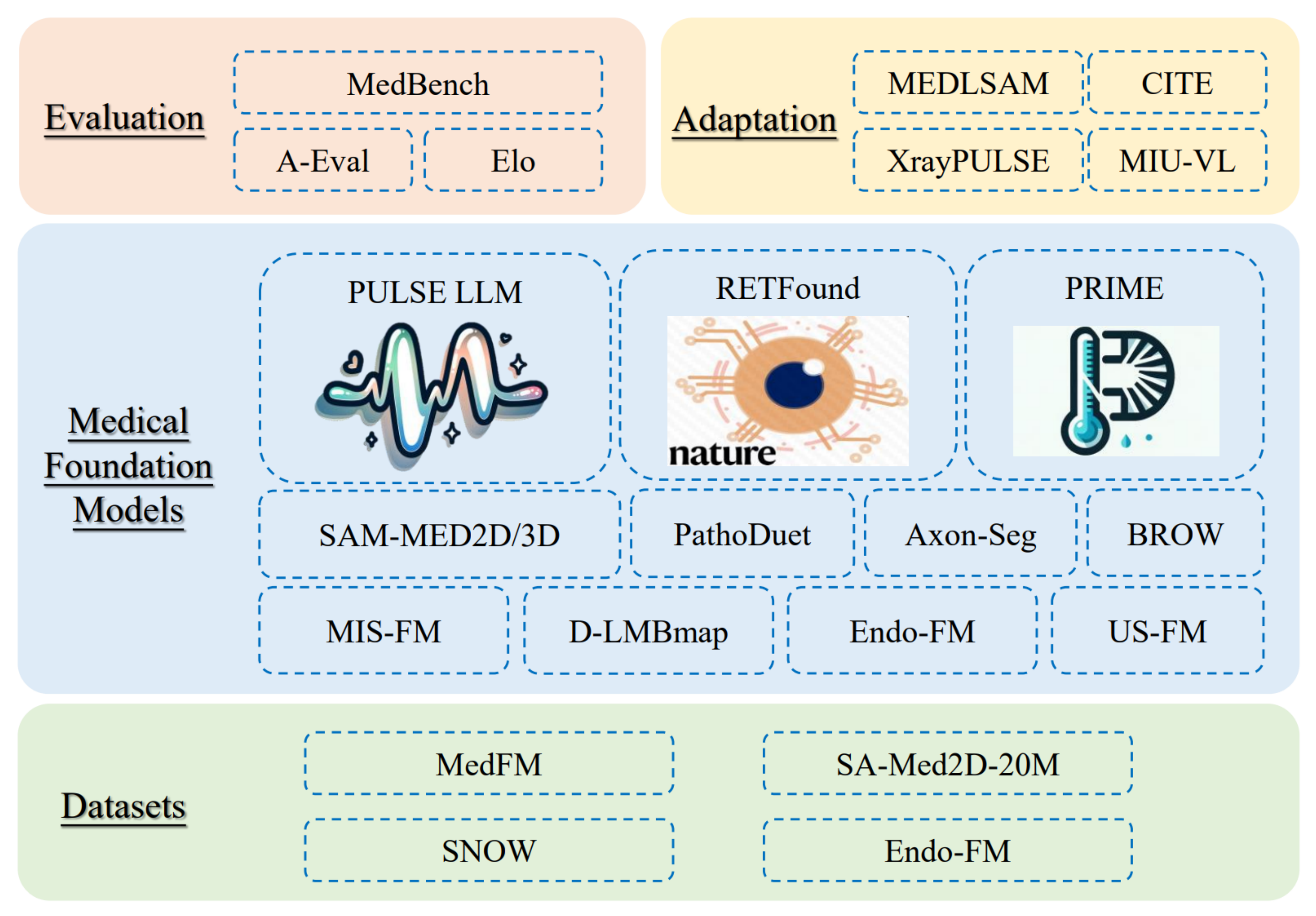}}
	\caption{The overall organization of OpenMEDLab \cite{Wang2024-ec}.}
	\label{fig:openmedlab}
	%\Description{Pipelines of robust watermarking, fragile watermarking, and our SepMark.}
\end{figure}

\section{Challenges}
Technical limitations of large medical models arise from inadequate high-quality training datasets, posing challenges in ensuring the accuracy and effectiveness of generated information. High-quality medical data and accurate annotations are essential for model training and application \cite{vamathevan2019applications}. Inaccurate or incomplete data annotation can impact model performance and stability, leading to inaccurate output answers. Furthermore, given the ongoing research and innovation in the medical field, outdated content can exacerbate inaccuracies in results. The most concerning issue associated with the use of AI large models is AI hallucination, where outputs are generated that sound plausible but are incorrect or irrelevant to the context \cite{bajwa2021artificial}. Thus, radiologists require interpretability in AI large models to derive imaging recommendations and ensure the accurate delivery of information to patients and clinicians. Additionally, integrating AI large models into existing imaging workflows warrants consideration. Radiologists should have easy access to and innovative utilization of localized AI large models within current systems and frameworks. These models should promptly offer actionable insights and recommendations. Potential future implementations include integration into the PACS system as a "plug-in" or establishment as an independent text data processing center alongside PACS. Further research is required to ensure that large AI models alleviate workload rather than impose a greater usage burden on physicians.

Legal and ethical challenges associated with large models: Many universities globally have banned the use of large language models like ChatGPT for learning and examination tasks, and numerous publishers restrict ChatGPT's involvement as a collaborator in academic papers. Presently, the World Health Organization (WHO) advises exercising caution in deploying large artificial intelligence models like ChatGPT \cite{guidance2021ethics}. Unverified AI large models may lead to misdiagnoses in medical imaging, undermining public trust in such models. AI large models must address data security and privacy concerns. Governments, regulatory bodies, and medical institutions should enforce stringent data privacy protocols to safeguard data security and patient privacy. The design and application of AI large models should adhere to medical ethical principles, respect patients' rights and dignity, and safeguard patient privacy and safety. Doctors and researchers should adhere to medical ethics and professional obligations when employing large models to ensure the legality and ethical conduct of medical practices.

\section{Conclusion}
Artificial intelligence (AI) technology has rapidly advanced in the medical field due to improvements in computing power, computer hardware, and the emergence of the big data era, revolutionizing traditional medical practices. While providing convenience to clinical work, artificial intelligence is still in its early stages in the medical field as an emerging technology. Despite the vast potential of medical AI, current algorithm models in medicine must undergo further maturation. Consequently, additional research and improvements are necessary when implementing artificial intelligence technology to ensure its accuracy and reliability. Moreover, the safety of medical AI requires further enhancement. Given the importance of safeguarding patient privacy and data security, appropriate security measures must be implemented when promoting artificial intelligence technology. Through the application of artificial intelligence technology, we aim to address increasingly complex medical conditions in the future, turning the seemingly impossible into reality.

\section*{CRediT authorship contribution statement}
Liangrui Pan: Conceptualization, Methodology, Writing – original draft, Writing – review \& editing. Zhenyu Zhao : Conceptualization, Validation.  Ying Lu and Kewei Tang: Writing – review \& editing. Qingchun Liang and Shaoliang Peng: Supervision.

\section*{Declaration of competing interest}
There are no funds and conflict of interest available for this manuscript.

\section*{Acknowledgment}
\begin{sloppypar}
This work was supported by NSFC-FDCT Grants 62361166662; National Key R\&D Program of China 2023YFC3503400, 2022YFC3400400; Key R\&D Program of Hunan Province 2023GK2004, 2023SK2059, 2023SK2060; Top 10 Technical Key Project in Hunan Province 2023GK1010; Key Technologies R\&D Program of Guangdong Province (2023B1111030004 to FFH). The Funds of State Key Laboratory of Chemo/Biosensing and Chemometrics, the National Supercomputing Center in Changsha (\url{http://nscc.hnu.edu.cn/}), and Peng Cheng Lab.

\end{sloppypar}

%\section*{References}

%\bibliographystyle{SELECTOR}
%\bibliography{SELECTOR}
\bibliographystyle{unsrt}

\bibliography{REFERENCES1}

\end{document}